# Admittance Controller Complemented with Real-time Singularity Avoidance for Rehabilitation Parallel Robots

José L. Pulloquinga[a,*], Rafael J. Escarabajal[a], Marina Vallés[a], Miguel Díaz-Rodríguez[c], Vicente Mata[b], Ángel Valera[a]

[a]*Departamento de Ingeniería de Sistemas y Automatica, Universitat Politècnica de València, Camino de Vera, s/n, Valencia, 46022, Valencia, Spain*
[b]*Departamento de Ingeniería Mecánica y de Materiales, Universitat Politècnica de València, Camino de Vera, s/n, Valencia, 46022, Valencia, Spain*
[c]*Departamento de Tecnología y Diseño, Escuela de Ingeniería Mecánica, Universidad de los Andes, Núcleo la Hechicera, Merida, 5101, Merida, Venezuela*

**Abstract**

Rehabilitation tasks demand robust and accurate trajectory-tracking performance, mainly achieved with parallel robots. In this field, limiting the value of the force exerted on the patient is crucial, especially when an injured limb is involved. In human-robot interaction studies, the admittance controller modifies the location of the robot according to the user efforts driving the end-effector to an arbitrary location within the workspace. However, a parallel robot has singularities within the workspace, making implementing a conventional admittance controller unsafe. Thus, this study proposes an admittance controller that overcomes the limitations of singular configurations by using a real-time singularity avoidance algorithm. The singularity avoidance algorithm modifies the original trajectory based on the actual location of the parallel robot. The complemented admittance controller is applied to a 4 degrees of freedom parallel robot for knee rehabilitation. In this case, the actual location is measured by a 3D tracking system because the location calculated by the forward kinematics is inaccurate in the vicinity of a singularity. The experimental results verify the effectiveness of the proposed admittance controller for safe knee rehabilitation exercises.

## 1. Introduction

Robotic rehabilitation can help patients recover and improve their mobility by performing repetitive movements aided by a robot under the supervision of a physiotherapist [1]. For example, a robot can move the human limb accurately to perform lower limb rehabilitation over prolonged sessions [2]. Among rehabilitation robots, a parallel robot (PR) offers excellent load capacity, stiffness, and accuracy [3]. Thus, these features allow integrating PRs into rehabilitation tasks [4]. In general, PRs are mechanical devices that control the end-effector by at least two closed kinematic chains where only a subset of joints are actuated [5, 6].

Motion rehabilitation therapies can be divided into patient-passive and patient-active exercises. In the patient-passive exercise, the robot follows a reference trajectory defined by the therapist without considering the patient interaction, i.e., the robot requires a trajectory controller [7]. In a patient-active exercise, the robot modifies the motion defined by the therapist according to the forces exerted by the patient [8]. The control strategies applied to patient-active exercises are admittance control [9], hybrid force/position control [10], bio-signals based control [11], and adaptive control [12]. The admittance control allows a dynamic relationship between the robot position and the patient effort making it one of the most appropriate for rehabilitation [8].

The admittance controller interacts with the patient by modifying the original trajectory according to the force exerted by the limb during rehabilitation [13]. The original trajectory can be modified using a mass-spring-damper system that provides compliant behaviour to the robot within the entire workspace [14, 15]. However, singular configurations, such as Type II singularity, arise inside the workspace of a non-redundant PR [16]. In this type of singularity, the PR cannot withstand external forces even if the actuators are locked, i.e., the patient loses control of the end-effector [17]. Thus, conventional admittance controllers cannot be straightforwardly implemented in a PR because the patient could unintentionally drive the PR to a Type II singularity, which is an unsafe position.

McDaid, Tsoi and Xie [18] applied an admittance controller to an ankle rehabilitation PR after optimising the singularity-free workspace. Dong et al. [19] implemented an admittance controller by constraining the workspace by limiting all degrees of freedom (DOF) of the PR using mechanical stoppers. However, the optimisation procedures reduce the workspace of the PR, and in most non-redundant PR, a small percentage of singularities remain within the workspace [20]. As the PR interacts with an injured limb or with a constrained mobility limb, ensuring an accurate avoidance of Type II singularities is essential. Thus, an additional approach is required to handle Type II singularities to make the admittance controller suitable for rehabilitation tasks with a PR.

Pulloquinga et al. [21] proposed the minimum angle be-tween two instantaneous screw axes from the Output Twist Screws ($\Omega_{i,j}$) to measure the closeness to a Type II singularity in spatial PRs. The minimum angle $\Omega_{i,j}$ is calculated based on the actual location and orientation (pose) of the PR where the subindices $i, j$ identify the actuators involved in the singular configuration. The study showed that the minimum $\Omega_{i,j}$ is an effective proximity detector for Type II singularities. However, the approach was not implemented for offline or online trajectory planning.

*Corresponding author
Email addresses:* jopulza@doctor.upv.es (José L. Pulloquinga), raessan2@doctor.upv.es (Rafael J. Escarabajal), mvalles@isa.upv.es (Marina Vallés), dmiguel@ula.ve (Miguel Díaz-Rodríguez), vmata@mcm.upv.es (Vicente Mata), giuprog@isa.upv.es (Ángel Valera)

Subsequently, in [22], the authors developed a vision-based controller to release a 4-DOF PR for knee rehabilitation from a Type II singularity. The proposed controller releases the PR from a singularity by modifying the trajectory of the pair of actuators identified by the minimum $\Omega_{i,j}$. The actual pose of the PR is measured by a 3D tracking sensor to deal with the non-unique solution of solving the forward kinematics near a Type II singularity [23]. This approach cannot prevent the 4-DOF PR from reaching a singular configuration. Thus, in [24] a Type II singularity avoidance algorithm based on the minimum $\Omega_{i,j}$ was developed for offline singularity-free trajectory planning in a 4-DOF PR. In [25] the proposed Type II singularity avoidance algorithm was extended for online singularity-free trajectory planning in a 4-DOF PR for knee rehabilitation. The proposed singularity avoidance algorithm exhibits low computational cost and minimum deviation of the original trajectory. However, the Type II singularity avoidance algorithm was not applied to patient-active knee rehabilitation with a human limb, i.e., the proposed algorithm was not involved in force/position control.

This study proposes an admittance controller improved with a real-time singularity avoidance algorithm for PRs. The proposed controller is applied to a 4-DOF PR for knee rehabilitation tasks. The contributions of this study are twofold. The first is to improve the admittance control with a real-time singularity avoidance algorithm. The complemented admittance controller ensures complete control over the end-effector during human-robot interaction without optimising the workspace or adding mechanical limits. The effectiveness of the complemented admittance controller in rehabilitation procedures is verified by executing patient-active exercises with the actual 4-DOF PR. The patient-active exercises are performed with a mannequin limb and a human limb. The second contribution is that the proposed Type II singularity avoidance algorithm in [25] was combined with a trajectory controller. Therefore, the singularity avoidance algorithm is combined for the first time with a force/position control law.

The remaining of this manuscript is as follows. Section 2 introduces the 4-DOF PR for knee rehabilitation that has been used for the experiments, the 3D vision system employed to measure the pose of the PR, and the force sensor characteristics. Section 3 describes the singularities in PRs, the conventional admittance controller, the algorithm to avoid Type II singularities, and the proposed complemented admittance controller. Finally, the experimental results on the actual PR and the conclusions are presented in Section 4 and Section 5, respectively.

## 2. Equipment and materials

### 2.1. Knee rehabilitation parallel robot

The goal of the knee rehabilitation process is to recover the plasticity of the lower limb on the knee's ligaments after a knee injury or surgery. The knee rehabilitation procedure requires three fundamental movements of the lower limb: flexion-extension of the knee, flexion of the hip and internal-external knee rotation [26].

The human leg requires two independent translations in the tibiofemoral plane combined with one autonomous rotation of the ankle for the flexion-extension of the knee. The internal-external knee rotation requires that the human leg rotates around the coronal plane. Thus, a PR for knee rehabilitation requires four DOFs, two translations and two rotations. A 3U$\underline{P}$S+R$\underline{P}$U PR provides the four DOFs required for the fundamental knee rehabilitation movements [27], see Fig. 1a. This PR has three external limbs with UPS configuration and a central limb with RPU configuration. The letters P, R, S, and U stand for prismatic, revolute, spherical, and universal joints, respectively, where the underlined letters represent the actuated joints.

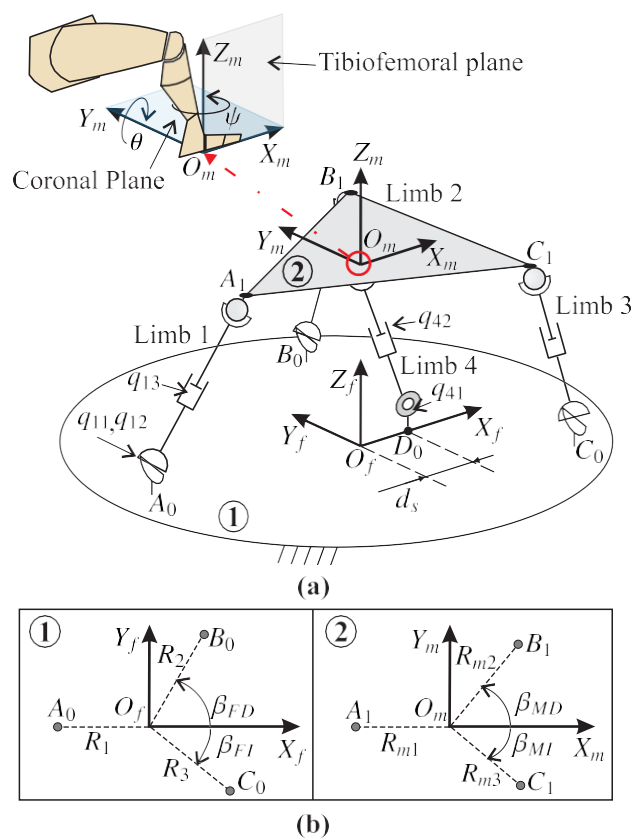

Figure 1: Knee rehabilitation PR (a) simplified view (b) configuration in 1. fixed platform and 2. mobile platform.

In contrast to other 4-DOF PRs, the 3U$\underline{P}$S+R$\underline{P}$U PR is suitable for knee rehabilitation because it provides a high payload capacity with a compact size [28]. Moreover, the revolute and universal joints in the central limb supply invariant constraints of the translation in the $Y_m$ axis and the rotation around the $X_m$ axis (Fig. 1a).

The pose of the knee rehabilitation PR is represented by the $\vec{X}$ as follows:

$$\vec{X} = \begin{bmatrix} x_m & z_m & \theta & \psi \end{bmatrix}^T \quad (1)$$

where $x_m$ and $z_m$ represent the two translational movements in the tibiofemoral plane, while $\theta$ and $\psi$ stand for the rotation around the $Y_m$ and $Z_m$ axes, respectively.

The $\vec{X}$ is controlled by four linear actuators represented by $\vec{q}_{ind}$, defined as:

$$\vec{q}_{ind} = \begin{bmatrix} q_{13} & q_{23} & q_{33} & q_{42} \end{bmatrix}^T \quad (2)$$

with $q_{13}$, $q_{23}$, $q_{33}$ and $q_{42}$ are the length of the linear actuators on the limbs 1,2,3 and 4 respectively.

The orientation of the external limbs is defined by the universal joints represented by the variables $q_{l1}$, $q_{l2}$ where $l = 1 \ldots 3$. The orientation of the central limb is defined by the revolute joint $q_{41}$.

The configuration of the PR is defined by points $A_0$, $B_0$, $C_0$, $D_0$ connecting the four limbs to the fixed platform, and $A_1$, $B_1$, $C_1$, $O_m$ connecting the limbs to the mobile platform, see Fig. 1a. In Fig. 1b, points $A_0$, $B_0$, $C_0$, $D_0$ are defined by $R_1$, $R_2$, $R_3$, $\beta_{FD}$, $\beta_{FI}$, and $d_s$ measured with respect to $O_f X_f Y_f Z_f$. The locations of $A_1$, $B_1$, $C_1$, and $O_m$ are defined by the geometric variables $R_{m1}$, $R_{m2}$, $R_{m3}$, $\beta_{MD}$, and $\beta_{MI}$ measured with respect to the mobile reference system $O_m X_m Y_m Z_m$.

In this study, the PR was configured with $R_1 = R_2 = R_3 = 0.3\ m$, $\beta_{FD} = 5\ °$, $\beta_{FI} = 90\ °$, $d_s = 0.0\ m$, $R_{m1} = R_{m2} = R_{m3} = 0.2\ m$, $\beta_{MD} = 70\ °$, $\beta_{MI} = 30\ °$.

*2.2. Vision-based tracking sensor*

A 3D tracking system (3DTS) was implemented as a vision-based sensor to track the location and orientation of the knee rehabilitation PR. The 3DTS allows an accurate avoidance of Type II singularities, increasing the safety of the patient. In particular, the 3DTS consists of 10 infrared cameras (Flex 13) manufactured by OptiTrack, see Fig. 2. The cameras can achieve an average accuracy greater than 0.1 mm. Moreover, the cameras have a resolution of 1.3 Megapixels at 120 Hz, i.e., 8.3 ms between two subsequent captures.

In the architecture of the 3DTS (see Fig. 3a), the cameras are connected to two OptiHub2 devices to record the captured images. The OptiHub2 devices send the image data through a high-speed USB to the camera control computer. Finally, the location data are sent to the PR control computer via an Ethernet connection.

In the camera control computer, the software Motive, provided by OptiTrack, processes the 2D camera images in the 3D location of markers inside the tracking area. In particular, the software Motive can associate a custom set of markers in a virtual object defined as a rigid body that can be set with 3 to 20 markers. A marker is a sphere covered with a reflective material. Motive can perform a real-time stream of the location and orientation of all rigid bodies, including the location of each marker, via Ethernet using unicast or multicast protocols.

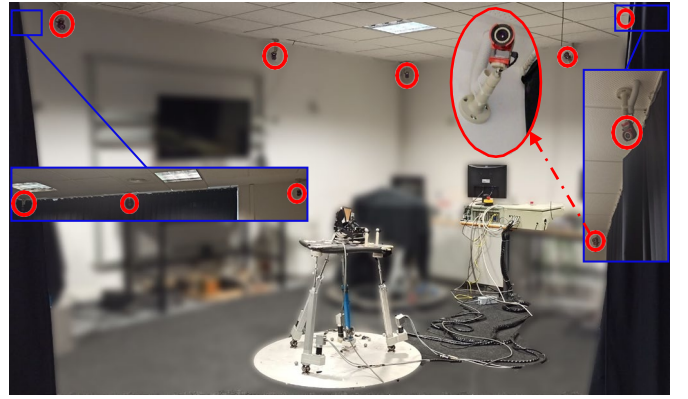

Figure 2: Overall view of the OptiTrack 3DTS.

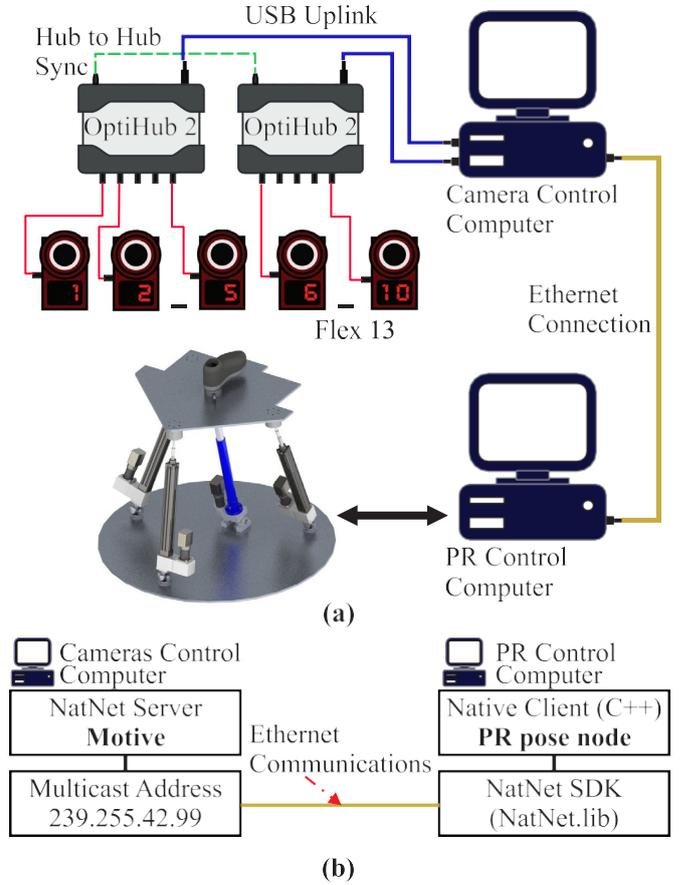

Figure 3: Architecture of OptiTrack 3DTS (a) hardware (b) software.

The streaming data are accessed by a client/server architecture based on the NatNet software development kit (NatNet SDK).

This study implemented a NatNet client/server architecture where the server could be run in the camera control computer, and a native client could be executed in the PR control computer (see Fig. 3b). The native client had attached to a C++ handler named **PR pose node** that was executed when a new frame of data was available.

The actual pose of the 3U<u>P</u>S+R<u>P</u>U PR was retrieved by measuring the 3D location of six markers. Three markers were placed on the mobile platform while the other three were attached to the fixed platform, where both sets were elements of two different rigid bodies. Given the location of the six markers, the actual position and orientation of the mobile platform were calculated with respect to the fixed frame $O_f X_f Y_f Z_f$. The actual position and orientation of the mobile platform were represented by $\vec{X}_c$ and sorted as $[x_m\ z_m\ \theta\ \psi]^T$, analogous to expression (1).

All cameras required calibration to ensure a correct reconstruction of the 3D location provided by the 3DTS. The calibration process is:

1. Set the orientation of the cameras so that they focus on the centre of the tracking area.
2. Adjust the brightness and illumination of the cameras to avoid the detection of unwanted objects.
3. Adjust the relative position of the cameras by moving the calibration wand provided by OptiTrack. An empty tracking area is required.
4. Set the ground plane for the tracking area with a calibration square provided by OptiTrack.

Steps 2-4 of the calibration process were executed in Motive, consuming less than five minutes.

### 2.3. Force/torque sensor

The admittance control requires an accurate measurement of the external force applied to the PR. The proposed system measures the forces and moments exerted by the patient's foot on the mobile platform using an FTN-Delta sensor. The FTN-Deltas sensor is a six-axis force/torque sensor (Schunk). For the knee rehabilitation PR, the FTN-Delta sensor was installed at $O_m$ on the mobile platform, see Fig. 4a. The forces measured in the $x$, $y$, and $z$ axes of the sensor reference frame can be represented by $F_x$, $F_y$, and $F_z$, respectively. Similarly, $M_x$, $M_y$, and $M_z$ stand for the moments measured by the FTN-Delta sensor in the reference frame $\{O-xyz\}$, see Fig. 4a. The FTN-Delta sensor measures force with a resolution of 0.065 N for $F_x$, $F_y$, and 0.125 N for $F_z$. The moments $M_x$, $M_y$, and $M_z$ are measured with a resolution of 0.004 N m. The FTN-Delta sensor has a measuring range of $\pm$ 330 N for $F_x$, $F_y$ and $\pm$990 N for $F_z$. The measuring range for the moments $M_x$, $M_y$, and $M_z$ is $\pm$ 30 N·m. The data from the FTN-Delta sensor were amplified, filtered, and transmitted by a Netbox NETB that supports CAN bus, DeviceNet, PROFINET, and UDP interface protocols at up to 7 Mhz.

The Netbox NETB was connected to the PR control computer as a single client using a UDP interface on the Ethernet protocol, see Fig. 4a. The single client had a C++ handler named **Force node** attached, which was executed when the admittance controller required a new frame of data, see Fig. 4b. The force node provided the measurements of force and moments in the 4-DOF of the knee rehabilitation PR in the vector $\vec{F}_c$ as follows:

$$\vec{F}_c = \begin{bmatrix} F_x & F_z & M_y & M_z \end{bmatrix}^T \quad (3)$$

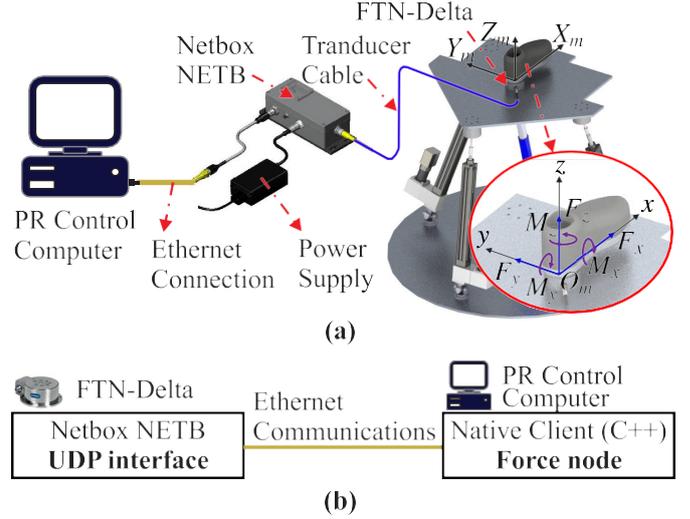

Figure 4: Architecture of FTN-Delta sensor (a) hardware (b) software.

where the $F_x$ stands for the force exerted by the patient on the axis $X_m$, and the $F_z$ is the force exerted on the axis $Z_m$. The $M_y$ and $M_z$ are the moments exerted by the patient around the tibiofemoral and coronal planes, respectively. The reference frame of the FTN-Delta sensor matches the mobile reference frame, see Fig. 4a.

The zero adjustments of forces and moments were performed considering no load on the mobile platform. Moreover, a death zone was defined as three times the standard deviation of data measured with the sensor unloaded to account for the noise.

## 3. Proposed Controller

First, the singularities in a PR and the admittance control law are described. Subsequently, the method to detect the proximity to a Type II singularity based on Output Twist Screws is explained. After that, the proposed algorithm to avoid a Type II singularity is explained. Finally, the admittance controller complemented with the Type II singularity avoidance algorithm is shown.

### 3.1. Singularities in parallel robots

In [16], Gosselin and Angeles define the velocity relationship between the cartesian coordinates and joint coordinates as follows:

$$J_D \vec{\dot{X}} + J_I \vec{\dot{q}}_{ind} = \vec{0} \quad (4)$$

where $\vec{\dot{X}}$ stands for the velocity of the end-effector, $\vec{\dot{q}}_{ind}$ represents the velocity of the actuators, $J_D$ is the forward Jacobian matrix, and $J_I$ is the inverse Jacobian matrix. For a non-redundant PR, $J_D$ and $J_I$ are square ($df\ x\ df$), with $df$ as the number of the DOFs of the PR.

According to (4), the rank deficiency of the Jacobian matrices defines two types of singularities. In a Type I singularity ($\|J_I\| = 0$), the PR loses mobility in at least one direction, i.e., the PR reaches the workspace boundary. Thus, Type I singularities are not a challenging task [29]. In a Type II singularity ($|J_D| = 0$), the PR loses control of the end-effector at least in one direction despite all actuators being locked. In this singular configuration, if an external force is applied to the end-effector, an uncontrolled motion is produced at least in one arbitrary direction.

Type II singularities are configurations within the workspace where the PR loses the end-effector motion, making them potentially dangerous for the user and PR. In particular, rehabilitation robots interact with human limbs, where the robot must ensure complete control of the end-effector motion, making Type II singularities a major problem to solve.

*3.2. Admittance control*

In lower limb rehabilitation, a patient-active exercise lets a patient modify a predefined movement according to the pain limitation of the limb in rehabilitation [30]. Therefore, a robot must exhibit compliant behaviour according to the effort applied by the patient while a rehabilitation exercise is executed, i.e., force/position control is required. Compliant control uses the feedback force to modify directly (impedance control) or indirectly (admittance control) the dynamic behaviour of the controlled system [13]. The admittance controller has a cascade architecture, where the force control takes place in the outer loop, and the position control takes place in the inner loop [13]. The force controller aims to minimise the deviation in the force exerted by the patient ($\vec{e}_F$), which is defined as follows:

$$\vec{e}_F = \vec{F}_r - \vec{F}_c$$

where $\vec{F}_r$ represents the reference forces and moments for rehabilitation in the $df$ DOFs of the PR. $\vec{F}_c$ stands for forces and moments measured on the human limb in the same $df$ DOFs of the PR or configuration space.

The minimisation of $\vec{e}_F$ is achieved by modifying the reference trajectory for the location and orientation of the mobile platform or end-effector ($\vec{X}_r$) according to the expression:

$$\vec{X}_a = \vec{X}_r + \vec{\Delta X}$$

$\vec{X}_a$ stands for the trajectory generated in the outer loop. $\vec{\Delta X}$ is the modification on the pose of the PR to reach a specific $\vec{F}_r$. Both $\vec{X}_a$ and $\vec{\Delta X}$ are measured in configuration space. In the inner loop, the position controller takes $X_a$ as the set point.

Using the Laplace transform, the $\vec{\Delta X}$ is related to $\vec{e}_F$ by a second-order model as follows:

$$\vec{\Delta X}(s) = \vec{e}_F(s)(k + cs + ms^2)^{-1}$$

where $m$, $c$, and $k$ are square matrices defining the mass, viscous coefficient, and spring stiffness of the dynamic system, respectively.

In Fig. 5a, if $\vec{F}_c < \vec{F}_r$, the model in (7) calculates a $\vec{\Delta X}$ to move the PR against the human limb and increase the $\vec{F}_c$. When $\vec{F}_c > \vec{F}_r$, the $\vec{\Delta X}$ varies $\vec{X}_r$ to move the PR away from the limb under rehabilitation and reduce the $\vec{F}_c$, see Fig. 5b. Thus, the $\vec{\Delta X}$ enables the patient to drive the PR to an arbitrary pose within the workspace. However, if a Type II singularity arises, the PR loses control of the end-effector motion and could hurt the limb in rehabilitation. For example, during knee rehabilitation, if the patient flexes the knee because of pain and the PR extends the knee due to loss of control, that could damage the lower limb. Consequently, the conventional admittance controller is unsuitable for PR-assisted lower limb rehabilitation. Therefore, the conventional admittance controller must be improved using online detection and avoidance of Type II singularities.

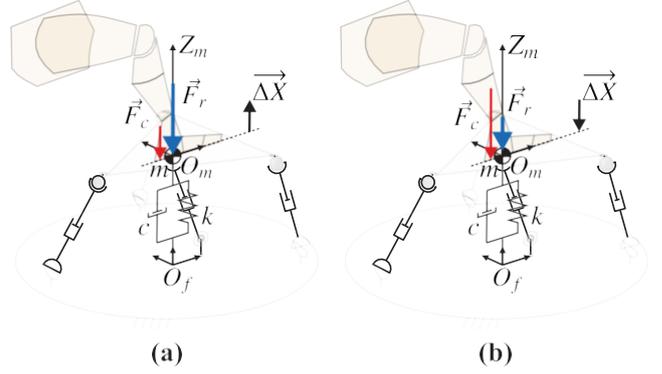

Figure 5: Compliant behaviour achieved by the second-order admittance model when (a) $\vec{F}_c < \vec{F}_r$ (b) $\vec{F}_c > \vec{F}_r$.

*3.3. Detection of proximity to a Type II singularity*

The motion of the mobile platform is produced by the combined action of $df$ actuators, i.e., the contribution of each actuator is challenging to identify. Using Screw Theory, Takeda and Funabashi [31] consider all actuators locked except one to divide the motion of the mobile platform ($) into $df$ Output Twist Screws (OTSs):

$$\$ = \rho_1 \hat{\$}_{O_1} + \rho_2 \hat{\$}_{O_2} + \ldots + \rho_{df} \hat{\$}_{O_{df}} \qquad (8)$$

where $\rho$ stands for the amplitude of each OTS, and $\hat{\$}_O$ represents the normalised OTSs.

If only one actuator can contribute to the motion of the mobile platform, a $\hat{\$}_O$ is defined by the work not applied by the locked actuators. Hereby, the $df$ $\hat{\$}_O$ are determined by solving the following equations:

$$\hat{\$}_{O_i} \circ \hat{\$}_{T_j} = 0 \quad (i, j = 1, 2, \ldots, df, \ i \ne j) \qquad (9)$$

with

$$\hat{\$}_O = \vec{\mu}_{\omega_O}; \vec{\mu}^*_{v_O} \qquad (10)$$

where $\circ$ is the reciprocal product, and $\hat{\$}_T$ stands for the normalised wrench screw transmitted by each actuator to the mobile platform. The $\vec{\mu}_{\omega_O}$ and $\vec{\mu}^*_{v_O}$ represent the angular and the instantaneous linear motion of the end-effector, respectively.

In a Type II singularity, at least two $\hat{\$}_O$s are linearly dependent [32], i.e., both $\vec{\mu}_{\omega_O}$ and $\vec{\mu}^*_{v_O}$ are parallel. This feature allows to measure the proximity to a Type II singularity using the angle between two different $\vec{\mu}_{\omega_O}$ named $\Omega_{i,j}$, as follows:

$$\Omega_{i,j} = \arccos(\vec{\mu}_{w_{O_i}} \cdot \vec{\mu}_{w_{O_j}}) \quad (11)$$

where $i$ and $j$ sub-indices are the two limbs under analysis, respectively. With $i, j = 1, 2, \ldots, df, i \neq j$.

The expression (11) comprises several indices $\Omega_{i,j}$. Therefore, the proximity detection to a Type II singularity is measured considering the minimum angle $\Omega_{i,j}$. The closeness to a Type II singularity is verified by the equality of the linear components $\vec{\mu}_{v_O}^*$.

The capability of the minimum angle $\Omega_{i,j}$ to detect the proximity to a Type II singularity was verified experimentally [21]. A crucial feature of the minimum angle $\Omega_{i,j}$ is that the sub-indices $i$ and $j$ identify the limbs responsible for the singular configuration. In [22], the identification of the limbs responsible for the Type II singularity was applied to release the 3U$\underline{P}$S+R$\underline{P}$U PR from a singularity already reached. In this study, we proposed implementing the minimum angle $\Omega_{i,j}$ for an online avoidance algorithm to prevent the PR from reaching a Type II singularity. In particular, the Type II singularity avoidance algorithm complements the admittance control to ensure complete control within the workspace.

The instantaneous motion of the 3U$\underline{P}$S+R$\underline{P}$U PR is divided into $\hat{\$}_{O_1}$, $\hat{\$}_{O_2}$, $\hat{\$}_{O_3}$, and $\hat{\$}_{O_4}$. For this PR (11) defines six indices $\Omega_{1,2}$ $\Omega_{1,3}$, $\Omega_{1,4}$, $\Omega_{2,3}$, $\Omega_{2,4}$, and $\Omega_{3,4}$. In this case, the four $\hat{\$}_{O_1}, \ldots, a\hat{\$}_{O_4}$ are calculated by solving (9) as follows:

$$\hat{\$}_{T_1} = (\vec{z}_{12}; \vec{r}_{O_m A_1} \times \vec{z}_{12}) \quad \hat{\$}_{T_2} = (\vec{z}_{22}; \vec{r}_{O_m B_1} \times \vec{z}_{22})$$
$$\hat{\$}_{T_3} = (\vec{z}_{32}; \vec{r}_{O_m C_1} \times \vec{z}_{32}) \quad \hat{\$}_{T_4} = (\vec{z}_{41}; \vec{0}) \quad (12)$$

where $\vec{z}_{12}, \vec{z}_{22}, \vec{z}_{32}$, and $\vec{z}_{41}$ are the direction of the forces applied by each actuator with respect to the fixed frame. $\vec{r}_{O_m A_1}$, $\vec{r}_{O_m B_1}$, and $\vec{r}_{O_m C_1}$ stand for the location vector between the point $O_m$ and the vertices $A_1$, $B_1$ and $C_1$, respectively.

The $\vec{z}_{12}$ is defined by the universal joint orientation between limb 1 and the fixed platform (see Fig. 1a). The universal joint can be modelled by two revolute joints $q_{11}$ and $q_{12}$; thus, $\vec{z}_{12}$ is defined as follows:

$$\vec{z}_{12} = \begin{bmatrix} \cos q_{11} \sin q_{12} & -\cos q_{12} & \sin q_{11} \sin q_{12} \end{bmatrix}^T \quad (13)$$

The $\vec{z}_{22}$ is calculated analogously to $\vec{z}_{12}$ by using $q_{21}$ and $q_{22}$ instead of $q_{11}$ and $q_{12}$. The same procedure is applied to calculate $\vec{z}_{32}$. In the central limb, the $\vec{z}_{41}$ is defined by the orientation of the revolute joint $q_{41}$ as:

$$\vec{z}_{41} = \begin{bmatrix} -\sin q_{41} & 0 & \cos q_{41} \end{bmatrix}^T \quad (14)$$

For a detailed explanation of the calculation of the six indices $\Omega_{1,2}, \Omega_{1,3}, \ldots, \Omega_{3,4}$ the reader could review [21].

### 3.4. Type II singularity avoidance algorithm

In the outer loop, the Type II singularity avoidance algorithm considers the reference $\vec{X}_a$ to generate a singularity-free trajectory for the actuators $\vec{q}_{ind_d}$ based on measuring $\vec{X}_c$. The inputs $\vec{X}_a$ and $\vec{X}_c$ are the references generated by the admittance control and the actual pose reached by the PR, respectively. A general overview of the proposed avoidance algorithm is shown in Fig. 6.

First, the Type II singularity avoidance algorithm calculates all possible indices $\Omega_{i,j}$ based on the $\vec{X}_a$ stored at $\vec{V\Omega}_a$. Next, the minimum element in $\vec{V\Omega}_a$ is calculated in $min\Omega_a$. In parallel, the $\vec{V\Omega}_c$ and $min\Omega_c$ are calculated for $\vec{X}_c$. The $\vec{i}_{av}$ stores the two limbs responsible for the Type II singularity identified by $min\Omega_c$. Finally, the proposed avoidance algorithm calculates the $\vec{q}_{ind_d}$ as follows:

$$\vec{q}_{ind_d} = \vec{q}_{ind_a} + v_d t_s \vec{\Delta \iota} \quad (15)$$

where $\vec{q}_{ind_a}$ stands for the joint space trajectory for $\vec{X}_a$ calculated by the inverse kinematics. The $v_d$ is the avoidance velocity for each actuator, and $t_s$ is the controller sample time. The $\vec{\Delta \iota}$ stands for the integer deviation required in the $df$ actuators to avoid a Type II singularity.

If $min\Omega_a$ is below a predefined threshold $\Omega_{lim}$, the $\vec{\Delta \iota}$ is modified as the reference $\vec{X}_a$ is approaching to a Type II singularity. The $\vec{\Delta \iota}$ is modified by one in the two rows corresponding to the elements in $\vec{i}_{av}$. The two rows identified by $\vec{i}_{av}$ could hold, increase, or decrease (the actuator stops or moves forward or backward), generating eight possible modifications of $\vec{\Delta \iota}$. The best option from the eight modifications of $\vec{\Delta \iota}$ is selected to generate a $\vec{q}_{ind_d}$ within the workspace and increase the value of $min\Omega_c$. The reference $\vec{q}_{ind_a}$ is continuously modified in Type II singularity avoidance, the output $ext_{pin}$ deactivates the successive iterations until $X_c$ becomes non-singular. During the admittance controller to prevent the patient from driving the PR back to a singular configuration. As soon as the reference $\vec{X}_a$ becomes non-singular, the binary variable $ext_{pin}$ is reactivated.

After avoiding a Type II singularity, $min\Omega_c \geq \Omega_{lim}$, if the reference $\vec{X}_a$ becomes non-singular ($min\Omega_a \geq \Omega_{lim}$), $\vec{\Delta \iota}$ must return progressively to zero. The two rows of $\vec{\Delta \iota}$ with the maximum value ($max\Delta$) are identified. By unitary modifications in the rows of the $max\Delta$, all eight possible new $\vec{\Delta \iota}$ are calculated. The best option of $\vec{\Delta \iota}$ is selected to decrease the absolute value of $\vec{\Delta \iota}$ and generate a non-singular pose $\vec{q}_{ind_d}$. The returning procedure continues until $\vec{\Delta \iota} = 0$.

The proposed Type II singularity avoidance algorithm is suitable for real-time operation because no optimisation procedures are required. The accuracy during the avoidance algorithm depends on the parameters $t_s$, $v_d$, and $\Omega_{lim}$. Moreover, the parameter $t_s$ is defined by the admittance controller sample time, and the $v_d$ is set to the average velocity of the actuators on the PR under analysis. The $\Omega_{lim}$ is set experimentally, executing trajectories that progressively approach a Type II singularity.

The $\vec{X}_c$ could be measured using a 3DTS or solving forward kinematics based on embedded encoders on the actuators' joints. However, the forward kinematics could be inaccurate because the solution is not unique near a Type II singularity [23]. In rehabilitation, the PR interacts with an injured limb or with a constrained mobility limb, making it essential to ensure an accurate avoidance of Type II singularities. Thus, a 3DTS is preferred over the calculation based on the forward kinematics problem for robotic rehabilitation.

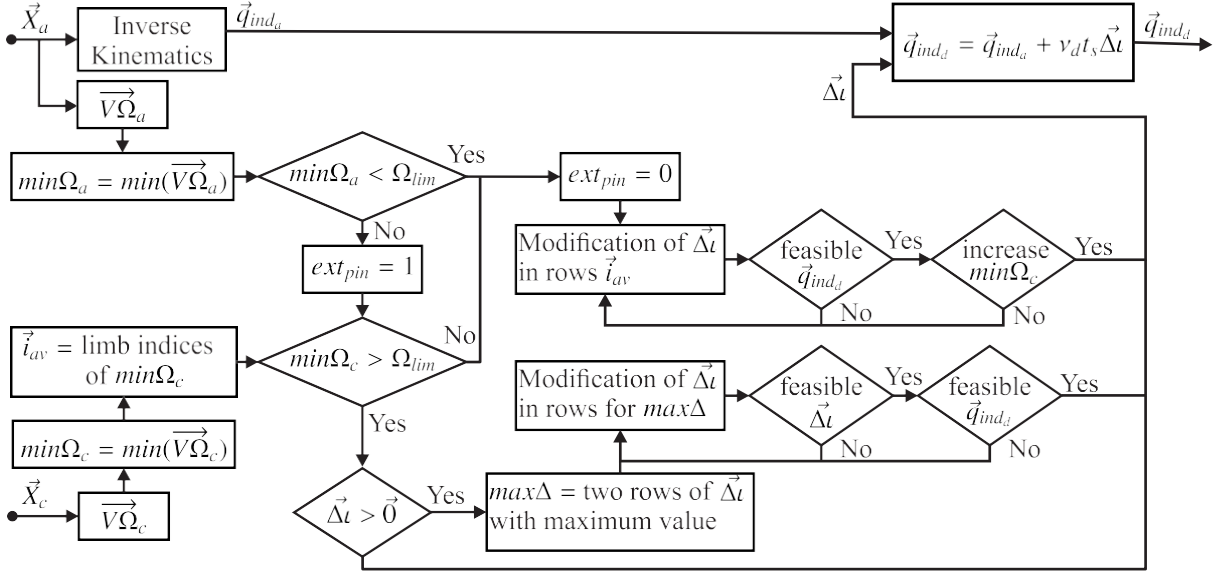

Figure 6: Diagram for Type II singularity avoidance algorithm.

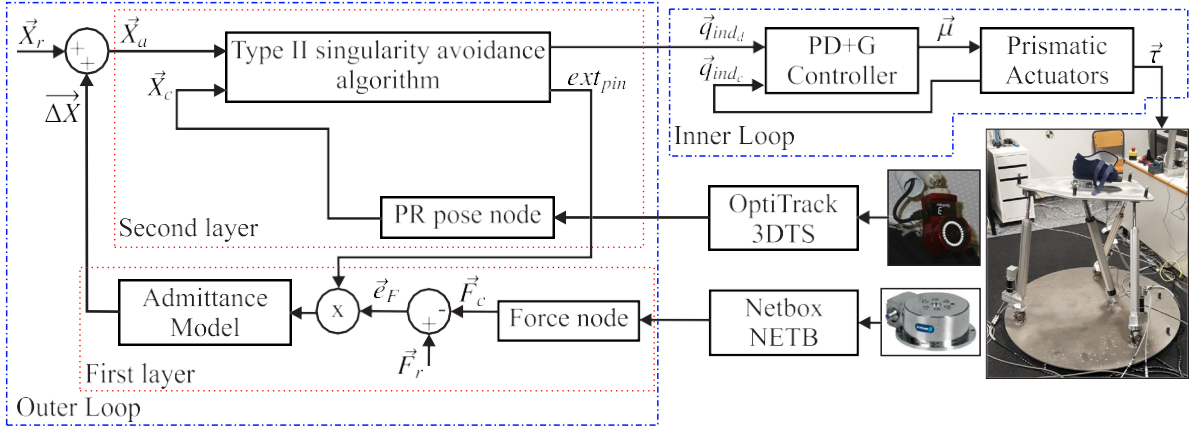

Figure 7: Architecture of the complemented admittance controller.

The Type II singularity avoidance algorithm developed for the 3U$\underline{P}$S+R$\underline{P}$U PR is described in Appendix A.

### 3.5. Complemented admittance controller description

The architecture of the admittance controller complemented with the Type II singularity avoidance algorithm for the 3U$\underline{P}$S+R$\underline{P}$U PR is shown in Fig. 7. Taking advantage of the cascade architecture of conventional admittance controllers, the outer loop is divided into two layers. In the first layer, an admittance model modifies the reference $\vec{X}_r$ by adding $\overrightarrow{\Delta X}$ to achieve the reference efforts $\vec{F}_r$ using the measurement $\vec{F}_c$ as feedback. In the second layer, the Type II singularity avoidance algorithm collects the adapted cartesian reference $\vec{X}_a$ and generates a singularity-free trajectory $\vec{q}_{ind_d}$ in the joint space. The avoidance algorithm keeps $\vec{q}_{ind_d} = \vec{q}_{ind_a}$ as long as $\vec{X}_a$ is not singular. The $\vec{q}_{ind_a}$ stands for the inverse kinematic solution of $\vec{X}_a$. If the reference $\vec{X}_a$ or the actual pose of the PR $\vec{X}_c$ is singular, the avoidance algorithm modifies $\vec{q}_{ind_a}$ to generate the singularity-free trajectory $\vec{q}_{ind_d}$. The proposed avoidance algorithm modifies the trajectory of the two actuators identified by the $min\Omega_c$. The $min\Omega_c$ is calculated based on the measurement of the actual pose of the PR $\vec{X}_c$.

When the reference $\vec{X}_r$ is non-singular, the trajectory $\vec{X}_a$ becomes singular due to the output $\overrightarrow{\Delta X}$ from the admittance model, see the outer loop in Fig. 7. Therefore, when avoiding a Type II singularity, the admittance model must temporally be deactivated to prevent conflict with the proposed avoidance algorithm. The admittance model could be deactivated by setting $\overrightarrow{\Delta X}$ to zero, see Section 3.2. However, if the output $\overrightarrow{\Delta X}$ is suddenly set to $\vec{0}$, the $\vec{X}_a$ presents random discontinuities that could hurt the human limb. In contrast, if the input $\vec{e}_F$ immediately changes to $\vec{0}$, the admittance model is gradually deactivated, i.e., the $\overrightarrow{\Delta X}$ decreases smoothly according to the dynamics of the second-order function (7). Therefore, the proposed

Type II singularity avoidance algorithm multiplies the binary variable $ext_{pin}$ by the $\vec{e}_F$ to deactivate the admittance model, the first layer of the outer loop in Fig. 7. In the proximity to a Type II singularity, the avoidance algorithm sets the $ext_{pin} = 0$ to decrease $\overrightarrow{\Delta X}$. As soon as the trajectory $\vec{X}_a$ becomes non-singular, the proposed avoidance controller sets $ext_{pin} = 1$ to reactivate the admittance controller. It is important to note that this mechanism enables the admittance control and the avoidance algorithm to coexist to ensure the safety of the limb under rehabilitation.

In the PR under analysis, the $\vec{F}_c$ is measured using the FTN-Delta sensor as described in Section 2.3. The 3DTS is used to determine $\vec{X}_c$. The system allows an accurate avoidance of a Type II singularity, increasing the safety of the PR during knee rehabilitation, see Section 2.2. The calculation of $\vec{X}_c$ by solving forward kinematics could be inaccurate near a Type II singularity. Hereby, the 3DTS is a solid option to ensure accurate singularity avoidance during a robotic rehabilitation procedure and ensure the integrity of the constrained mobility limb.

In the inner loop of the complemented admittance controller, a trajectory controller takes $\vec{q}_{ind_d}$ as a set point, see Fig. 7. In the PR under analysis, the trajectory controller is a proportional derivative controller with gravitational compensation (PD+G). The PD+G controller calculates the control actions ($\vec{\mu}$) required to track the trajectory $\vec{q}_{ind_d}$ based on the measurements of the location reached by the actuators $\vec{q}_{ind_c}$. The $\vec{\mu}$ are electrical signals proportional to the mechanical forces applied by the linear actuators $\vec{\tau}$.

The PD+G controller was selected because the 3UPS+RPU PR works at low velocity making unnecessary the compensation of the inertial, centrifugal, and Coriolis forces. The gravitational forces are calculated with high accuracy and low computing cost by using a model of eight base parameters identified in the actual PR [33]. Thus, the gravitational compensation (G) ensures a minimal steady-state error without the need for integral actions. However, the trajectory controller on the inner loop could be changed depending on the necessities of the application. Before selecting the PD+G controller, the authors implemented and analysed other control laws such as computed torque (CT), adaptive, linear algebra-based (LAB) and proportional integral derivative (PID).

Note that identifying a proper set of dynamic parameters of the admittance model for each user remains open. Nevertheless, this study aims to prove that Type II singular configurations in PRs can be overcome using admittance control by adding a real-time singularity avoidance algorithm.

## 4. Results and discussion

First, the experimental settings to test the proposed admittance controller complemented with the Type II singularity avoidance algorithm are presented. Subsequently, the performance of the complemented admittance controller is shown and compared with the conventional admittance controller.

### 4.1. Experimental setup

A patient-active exercise close to a Type II singularity was designed to compare the performance of the complemented admittance controller with that of the conventional admittance controller. The patient-active exercise comprised an internal-external knee rotation with the 3UPS+RPU PR. During the internal-external knee rotation, the patient had a zero-effort trajectory ($\vec{F}_r = \vec{0}$). Thus, the patient had complete control over the PR movements, i.e., the PR was completely compliant. The starting pose for the knee rotation exercise was set close to a Type II singularity by experimentally searching on the actual 3UPS+RPU PR based on the minimum angle $\Omega$. In this study, the reference $\vec{X}_r$ was constant and equal to the starting pose for the knee rotation exercise.

Before implementing the proposed complemented admittance controller on the actual PR, several simulations were executed using a virtual model of the human leg. Here, only the experiments with the actual robot are presented. The experiments have two phases designed to prioritise the patient's safety. Phase I considers a mannequin leg to mimic an internal-external knee rotation, see Fig. 8a. Phase II considered an uninjured and unrestricted human knee to perform the same internal-external knee rotation, see Fig. 8b. The experiment in Phase II was approved by the Ethics Committee of the Universitat Politècnica de València, and the patient signed an informed consent document.

The conventional admittance controller was applied to Phase I to expose the risk of the straightforward application to a PR. Phase II implemented the complemented admittance controller to verify that PR could overcome the singular configurations.

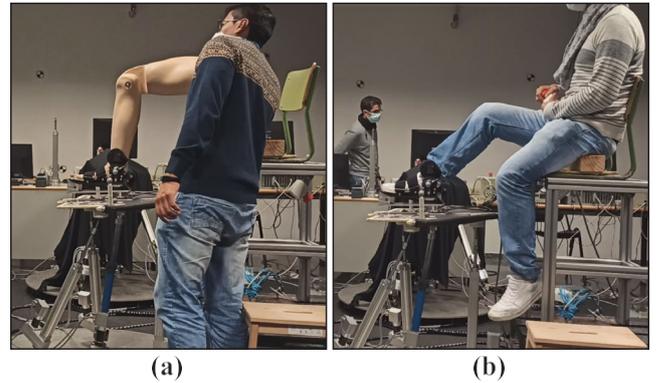

Figure 8: Experimental setup using (a) Mannequin (b) Human limb.

In this case, the admittance model was set experimentally to provide minimum resistance to the limb of the patient. Table 1 lists the parameters of the admittance controller. It is because our purpose was to verify that the proposed controller can overcome Type II singularities during admittance control. The identification of a suit set of $k$, $c$ and $m$ for knee rehabilitation was not discussed in this study.

During the flexion-extension of the knee, the latency of the lower limb muscular response exceeds 50 ms [34, 35]. Moreover, knee rehabilitation exercises are designed for low-velocity movement in impaired lower limbs. Thus, the proposed admittance controller is executed each 10 ms (100 Hz) to ensure a sufficiently fast response. The Type II avoidance algorithm, $v_d = 0.01$ *m/s* and $t_s = 0.01$ *s* to suit the maximum velocity during a knee rehabilitation procedure. After performing several experiments where the 3UPS+RPU PR moved to a Type II singularity, we set $\Omega_{lim} = 2°$ to prevent the loss of control on the mobile platform. The $\Omega_{lim}$ was set experimentally owing to the mechanical imperfections of the actual PR. For a profound explanation of the setting of $\Omega_{lim}$, the reader could refer to [21].

Table 1: Settings for the admittance model.

| Variable | Value |
|---|---|
| $k$ | $\begin{bmatrix} 250 & 0 & 0 & 0 \\ 0 & 500 & 0 & 0 \\ 0 & 0 & 25 & 0 \\ 0 & 0 & 0 & 25 \end{bmatrix}$ |
| $c$ | $\begin{bmatrix} 894 & 0 & 0 & 0 \\ 0 & 894 & 0 & 0 \\ 0 & 0 & 89.4 & 0 \\ 0 & 0 & 0 & 89.4 \end{bmatrix}$ |
| $m$ | $\begin{bmatrix} 200 & 0 & 0 & 0 \\ 0 & 200 & 0 & 0 \\ 0 & 0 & 20 & 0 \\ 0 & 0 & 0 & 20 \end{bmatrix}$ |

The proposed combined controller was executed modularly on an industrial computer using Robot Operating System 2 (ROS2) and the C++ programming language. The length of the actuators was measured by a PCI 1784 Advantech card connected to four quadruple AB phase encoder counters with a resolution of 500 counts per turn. The linear actuators were driven using four ESCON 50/5 servo controllers, where the $\vec{\mu}$ was sent through a 12-bit, 4-channel PCI 1720 Advantech card at a rate of 100 Hz.

### 4.2. Complemented admittance controller performance

In phase I, during the internal-external knee rotation, the conventional admittance controller drives the PR to a Type II singularity. In this case, the PR falls due to the gravity effect, and then the mannequin leg was extended abruptly. A recording of this experiment is available in $http://roboprop.ai2.upv.es/admittance\_without\_evader$.

Fig. 9 shows the tracking of PR location on the $Z_f$ axis, i.e., the height of the mobile platform during the knee rotation in phase I. In Fig. 9, the solid line $z_a$ represents the reference calculated by the conventional admittance controller, and the dashed line $z_c$ is the location measured in the actual PR. The $z_c$ shows the loss of control on the PR from $t = 11.46$ *s*, and the subsequent fall because the conventional admittance controller reaches a Type II singularity. Thus, the inherent danger of using a conventional admittance controller for human-robot interaction in rehabilitation based on PRs is verified.

Fig. 10 shows the proximity detection of the Type II singularity based on $min\Omega$. In this case, the proximity is detected by $\Omega_{2,4}$ during the entire experiment of the internal-external knee rotation. In the figure, $min\Omega_a$ represents the angle $\Omega_{2,4}$ for the reference calculated for the conventional admittance controller and $min\Omega_c$ is the index $\Omega_{2,4}$ for the actual pose reached by the PR. At $t = 11.46$ *s*, the $min\Omega_c < \Omega_{lim}$ verifies that the 3UPS+RPU PR reaches a Type II singularity, see Fig. 10. The index $min\Omega_c$ detects the proximity to a Type II singularity at $t = 9.77$ *s*, i.e., 1.69 *s* before the PR falls. Therefore, the $min\Omega$ is a feasible option to prevent the PR from reaching a Type II singularity.

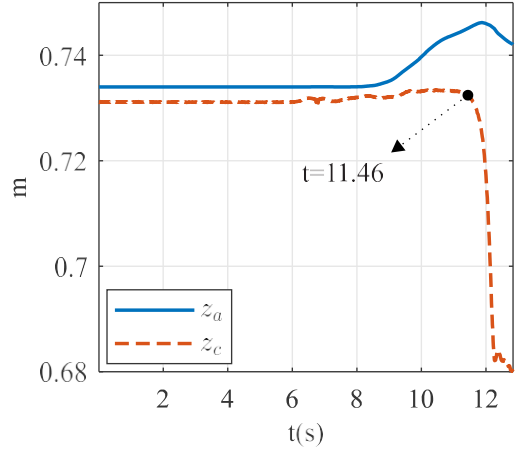

Figure 9: Conventional admittance controller: tracking location in $Z_f$.

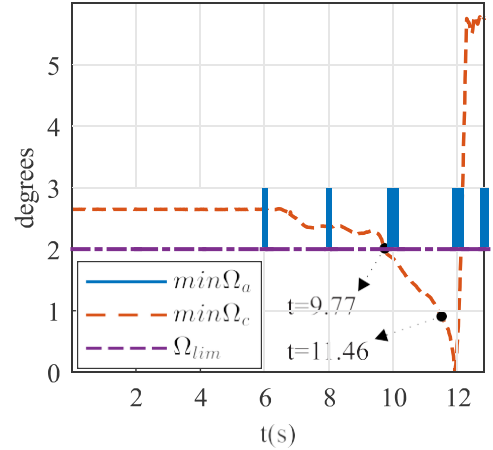

Figure 10: Conventional admittance controller: index $min\Omega$.

The admittance controller, complemented with the Type II singularity avoidance algorithm, effectively completes the two experimental phases without losing control. The execution of the complemented admittance controller during phase I and phase II are available in $http://roboprop.ai2.upv.es/admittance\_with\_evader\_vf/$ and $http://roboprop.ai2.upv.es/admittance\_with\_evader\_pierna\_humana\_vf/$, respectively.

Fig. 11 shows the proximity detection to a Type II singularity based on the $min\Omega$ for the complemented admittance controller during phase II. The $min\Omega_a$ represents the angle $\Omega_{2,4}$ calculated for the trajectory generated by the admittance model. The $min\Omega_c$ represents the angle $\Omega_{2,4}$ calculated for the pose measured on the actual PR. In Fig. 11, three periods arise where the patient tries to drive the PR to a Type II singularity, and the complemented controller avoids that singularity.

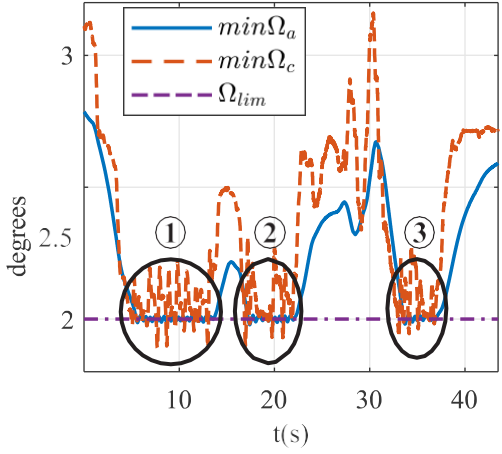

Figure 11: Complemented admittance controller: index $min\Omega$.

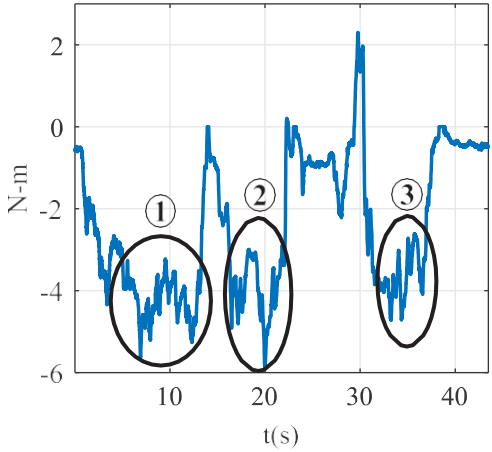

Figure 12: Complemented admittance controller: Moment around $Z_f$.

Fig. 12 shows the moment around the $Z_f$ measured during the internal-external knee rotation in phase II. Fig. 12 indicates that no discontinuities or sudden changes arise in the patient's foot during the three periods of Type II singularity avoidance. Note that during these three periods of Type II singularity avoidance, the patient did not perceive the deactivation of the admittance control. The non-perception of the deactivation of the admittance control was qualitatively verified by asking the patient. Therefore, the proposed complemented admittance controller overcomes the limitation of the Type II singularities in PRs, making it a safe option for knee rehabilitation.

The deviation from the prescribed trajectory due to Type II singularity avoidance was quantified using the mean absolute error (MAE) between the $\vec{q}_{ind_d}$ and $\vec{q}_{ind_a}$. This deviation could also be represented by the absolute percentage error (MAPE) between the same variables. If $\vec{q}_{ind_d}$ comes from the real-time Type II singularity avoidance algorithm and $\vec{q}_{ind_a}$ is generated by the admittance model, the MAE and MAPE are calculated as follows:

$$MAE = \frac{1}{df}\sum_{i=1}^{df}\left\{\sum_{k=1}^{n_t}\left(\frac{1}{n}\sum_{j=h}^{n}|q_{ind_a}(i,j) - q_{ind_c}(i,j)|\right)\right\} \quad (16)$$

$$MAPE = \frac{100}{df}\sum_{i=1}^{df}\left\{\sum_{k=1}^{n_t}\left(\frac{1}{n}\sum_{j=h}^{n}\left|\frac{q_{ind_a}(i,j) - q_{ind_c}(i,j)}{q_{ind_a}(i,j)}\right|\right)\right\} \quad (17)$$

where $n_t$ is the number of periods of activation of the singularity avoidance algorithm. $n$ is the number of samples in each $k$ period. $h$ is the initial instant of each period of activation of the singularity avoidance algorithm. $i$ and $j$ identify the actuator and the time instant analysed, respectively.

The absence of discontinuities or sudden changes in forces applied by the PR with the complemented admittance controller is verified by absolute variation rate (AVR). The AVR calculates the average instantaneous change between two subsequent time samples of the control actions during the singularity avoidance periods, i.e.:

$$AVR = \frac{1}{df}\sum_{i=1}^{df}\left\{\sum_{k=1}^{n_t}\left(\frac{1}{n}\sum_{j=h}^{n}|\tau(i,j) - \tau(i,j-1)|\right)\right\} \quad (18)$$

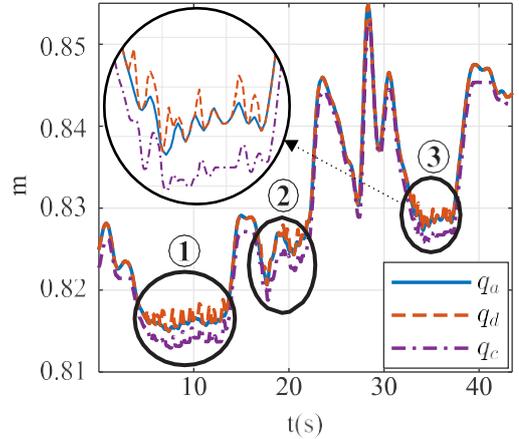

Figure 13: Complemented admittance controller: location tracking for the actuator on limb 2.

Table 2: Performance of the complemented admittance controller.

|  | Phase I | Phase II |
|---|---|---|
| MAE (mm) | 2.30 | 1.02 |
| MAPE (%) | 0.31 | 0.12 |
| AVR (N) | 3.42 | 3.83 |

Table 2 lists the MAE, MAPE, and AVR for the complemented admittance controller during the two experimental phases. This table shows that the deviation on the $q_{ind_a}$ during a Type

II singularity avoidance was lower than 2.5 mm (0.4%). The AVR during a Type II singularity avoidance is lower than 5 N. This variation in the forces exerted by the PR is negligible because during human walking the leg could exert 500 N on average [36]. Thus, the proposed combined admittance controller ensures a smooth behaviour of the 3UPS+RPU PR with a minimum deviation of the rehabilitation trajectory. In this paper, only the trajectories of the actuators on limbs 2 and 4 were modified as they are responsible for the singular configuration ($min\Omega_c = \Omega_{2,4}$). Fig. 13 shows the location of the actuator on limb 2 during the experiment in Phase II. The solid line $q_a$ represents the trajectory defined by the admittance model. The dashed line $q_d$ represents the trajectory generated by the real-time singularity avoidance algorithm, and $q_c$ is the actual location reached by the actuator. Fig. 13 indicates that the difference between $q_d$ and $q_a$ is less than 2.5 mm during the three periods of Type II singularity avoidance.

Ensuring the safety of patients with impaired limbs undergoing knee rehabilitation requires a high-accuracy robotic system. One crucial factor in achieving accuracy is the real-time Type II singularity avoidance algorithm, which relies on precise pose measurements of the PR. In this study, we utilised a 3DTS for accurate measurement of the PR pose. Although a 3DTS may be costly for a single PR, it can accurately measure the location of multiple PRs simultaneously, thereby increasing the productivity of rehabilitation sessions. For instance, a physiotherapist can supervise knee rehabilitation for a 4-DOF PR while another 3-DOF PR executes ankle rehabilitation.

## 5. Conclusions

This research evaluated the performance of a conventional admittance controller in a 3UPS+RPU PR designed for knee rehabilitation. The experimental results showed that this controller cannot maintain the stiffness required for the PR in the proximity of a Type II singularity, making it unsafe for human-robot interaction. This limitation was overcome by an admittance controller complemented with a real-time singularity avoidance algorithm. The proposed complemented admittance controller successfully provided complete control over the mobile platform during an internal-external knee rotation performed by an uninjured and unrestricted patient. The real-time singularity avoidance algorithm required a maximum deviation of 2.5 mm in the joint space trajectory to avoid a Type II singularity. The trajectory modification was introduced in the actuators involved in the singular configuration, identified by the $min\Omega_c$.

The proposed singularity avoidance algorithm, combined with the 3DTS, solves the limitations of conventional admittance control and provides a safe and effective option for rehabilitation based on PRs. While implementing a 3DTS for a single PR may be costly, it is able to track the pose of several PRs simultaneously, increasing the efficiency of the rehabilitation sessions. For non-hazardous applications, the pose of the PR can be estimated via forward kinematics using actuator feedback. Moreover, the avoidance algorithm remains the PR in non-singular poses, enabling the combination of an admittance model with a model-based controller without dynamic model degeneration.

In future work, the Type II singularity avoidance algorithm could be modified to smooth further the deviation of the actuators responsible for the singularity. Moreover, the complemented admittance controller has only been implemented in a 3UPS+RPU PR. The proposed controller will be tested in a 3RPS PR for ankle rehabilitation to verify the advantages provided by the admittance controller complemented with the real-time singularity avoidance algorithm. Finally, the 3UPS+RPU PR and the 3RPS PR could be analysed simultaneously to verify the advantages of using a 3DTS in robotic rehabilitation.

## Acknowledgements

This research was partially funded by Fondo Europeo de Desarrollo Regional (PID2021-125694OB-I00), cofounded by Programa de Ayudas de Investigación y Desarrollo de la Universitat Politècnica de València (PAID-01-19).

Funding for open access charge: CRUE-Universitat Politècnica de València

## Appendix A. Algorithm for avoiding Type II singularities

For the 3UPS+RPU PR, the complete process performed by the proposed Type II singularity avoidance algorithm is described in Algorithm 1. This algorithm verifies the feasibility of the possible modifications on the trajectory for the actuators by using Algorithm 2. The parameters of the Type II avoidance algorithm for the implementation on the 3UPS+RPU PR are shown in Table A.1. In this case, the vectors $\overrightarrow{maxq_{ind}}$ and $\overrightarrow{minq_{ind}}$ define the limits for the feasible location of the linear actuator of the PR. The vector $\vec{\alpha}_{lim}$ defines the maximum angle between the mobile platform and the linear actuator in the three external limbs, see Fig. 1a. The inputs required by the avoidance algorithm and the outputs calculated by the same algorithm is shown in Table A.2.

For this case, the elements of $\overrightarrow{V\dot{\Omega}}_a$ and $\overrightarrow{V\dot{\Omega}}_c$ are sorted as $\{\Omega_{1,2} \; \Omega_{1,3} \; \Omega_{1,4} \; \Omega_{2,3} \; \Omega_{2,4} \; \Omega_{3,4}\}^T$. The eight possible modifications of the two rows of the $\vec{\Delta \iota}$ are defined by the columns of the matrix $M_{av}$ as follows:

$$M_{av} = \begin{bmatrix} 1 & -1 & 1 & -1 & 1 & -1 & 0 & 0 \\ 1 & -1 & -1 & 1 & 0 & 0 & 1 & -1 \end{bmatrix} \quad (A.1)$$

The possible combinations of pair of actuators are defined by the rows in the matrix $M_{pa}$:

$$M_{pa} = \begin{bmatrix} 1 & 2 \\ 1 & 3 \\ 1 & 4 \\ 2 & 3 \\ 2 & 4 \\ 3 & 4 \end{bmatrix} \quad (A.2)$$

**Algorithm 1:** Type II singularity avoidance

**initialization**
| $\vec{\Delta\iota} = \vec{0}$
| $N_{av}$ = number of columns of $M_{av}$
**end**
**begin**
| $\vec{q}_{ind_d} = \vec{q}_{ind_a} + \nu_d t_s \vec{\Delta\iota}$
| Calculate $\vec{V\Omega}_a$ using $\vec{X}_a$
| Calculate $\vec{V\Omega}_c$ using $\vec{X}_c$
| $min\Omega_a$ = minimum element in $\vec{V\Omega}_a$
| $min\Omega_c$ = minimum element in $\vec{V\Omega}_c$
| $imin\Omega_c$ = row of minimum element in $\vec{V\Omega}_c$
| **if** $min\Omega_c < \Omega_{lim}$ **then**
| | $\vec{i}_{av}$ = row $imin\Omega_c$ from $M_{pa}$
| | $\vec{V\Omega}_{av}$ = $FnFeasibility(\vec{X}_c, \vec{q}_{ind_d}, M_{av}, N_{av}, \vec{i}_{av})$
| | $i_{max}$ = row of maximum element of $\vec{V\Omega}_{av}$
| | $\vec{\Delta\iota}(\vec{i}_{av}) = \vec{\Delta\iota}(\vec{i}_{av}) + M_{av}(:, i_{max})$
| **else if** $min\Omega_a > \Omega_{lim}$ AND $\vec{\Delta\iota} \neq \vec{0}$ **then**
| | $\vec{i}_{re}$ = first row from $M_{pa}$
| | $max\Delta$ = summation of $|\vec{\Delta\iota}(\vec{i}_{re})|$
| | **for** $k = 2$ **to** 6 **do**
| | | $\vec{i}_{aux}$ = row $k$ from $M_{pa}$
| | | $amax\Delta$ = summation of $|\vec{\Delta\iota}(\vec{i}_{aux})|$
| | | **if** $amax\Delta > max\Delta$ **then**
| | | | $max\Delta$ = $amax\Delta$
| | | | $\vec{i}_{re} = \vec{i}_{aux}$
| | | **end**
| | **end**
| | $c = 1$
| | **for** $k = 1$ **to** $N_{av}$ **do**
| | | $amax\Delta$ = summation of $|\vec{\Delta\iota}(\vec{i}_{re}) + M_{av}(:, k)|$
| | | **if** $amax\Delta < max\Delta$ **then**
| | | | $M_{re}(1:2, k) = M_{av}(:, c)$
| | | | $c$++
| | | **end**
| | **end**
| | $N_{re}$ = number of columns of $M_{re}$
| | $\vec{V\Omega}_{re}$ = $FnFeasibility(\vec{X}_c, \vec{q}_{ind_d}, M_{re}, N_{re}, \vec{i}_{re})$
| | $i_{max}$ = row of maximum element of $\vec{V\Omega}_{re}$
| | $\vec{\Delta\iota}(\vec{i}_{re}) = \vec{\Delta\iota}(\vec{i}_{re}) + M_{re}(:, i_{max})$
| **end**
| $\vec{q}_{ind_d} = \vec{q}_{ind_a} + \nu_d t_s \vec{\Delta\iota}$
| **if** $min\Omega_a > \Omega_{lim}$ **then**
| | $ext_{pin} = 1$
| **else**
| | $ext_{pin} = 0$
| **end**
**end**

**Algorithm 2:** Feasibility of all possible modifications.

**Fuction** $FnFeasibility(\vec{X}_c, \vec{q}_{ind_d}, M_{ch}, N_{ch}, \vec{i}_{ch})$:
| $\vec{V\Omega}_{ch}$ = zero vector of $N_{ch}$ rows
| **for** $k = 1$ **to** $N_{ch}$ **do**
| | $\vec{q}_{ch} = \vec{q}_{ind_d}$
| | $\vec{q}_{ch}(\vec{i}_{ch}) = \vec{q}_{ind_{ch}}(\vec{i}_{ch}) + \nu_d t_s M_{ch}(:, k)$
| | **if** $minq_{ind} < \vec{q}_{ch} < maxq_{ind}$ **then**
| | | $\vec{X}_{ch}$ = solve the forward kinematics for $\vec{q}_{ch}$, with $\vec{X}_c$ as initial condition
| | | $\vec{\alpha}_{ch}$ = Angle of spherical joints for $\vec{X}_{ch}$
| | | **if** $\vec{\alpha}_{ch} < \vec{\alpha}_{lim}$ **then**
| | | | $\vec{V\Omega}_{ch}(k)$ = calculate the index $\Omega_{i,j}$ for $\vec{X}_{ch}$ with $i, j = \vec{i}_{ch}$
| | | **end**
| | **end**
| **end**
| return $\vec{V\Omega}_{ch}$
**end**

Table A.1: Settings of the Type II singularity avoidance algorithm

| Variable | Default | Description |
|---|---|---|
| $\nu_d$ | 0.01 | avoidance velocity in $m/s$ |
| $t_s$ | 0.01 | controller sample time in $s$ |
| $\Omega_{lim}$ | 2 | experimental limit for $\Omega_{i,j}$ in degrees |
| $\overrightarrow{maxq_{ind}}$ | $\begin{bmatrix} 0.93 \\ 0.92 \\ 0.93 \\ 0.82 \end{bmatrix}$ | maximum feasible values for the actuators' length in $m$ |
| $\overrightarrow{minq_{ind}}$ | $\begin{bmatrix} 0.65 \\ 0.64 \\ 0.65 \end{bmatrix}$ | minimum feasible values for the actuators' length in $m$ |
| $\vec{\alpha}_{lim}$ | $\begin{bmatrix} 38 \\ 38 \\ 38 \end{bmatrix}$ | experimental limits for the spherical joints in degrees |

Table A.2: Inputs and Outputs of the Type II singularity avoidance algorithm.

**INPUTS**

| Variable | Description |
|---|---|
| $\vec{X}_a$ | reference for configuration space. Calculated by the oututer loop of the admittance controller |
| $\vec{X}_c$ | position and orientation measured by the 3DTS, feedback signal |

**OUTPUTS**

| | |
|---|---|
| $\vec{q}_{ind_d}$ | Singularity-free trajectory in joint space |
| $ext_{pin}$ | Indicator of non-singular $\vec{X}_r$ |

**References**

[1] S. Xie, Advanced Robotics for Medical Rehabilitation, 1st Edition, Vol. 108, Springer International Publishing, 2016. doi:10.1007/978-3-3

19-19896-5.
URL http://link.springer.com/10.1007/978-3-319-19896-5

[2] M. Zhang, T. Davies, S. Xie, Effectiveness of robot-assisted therapy on ankle rehabilitation – a systematic review, Journal of NeuroEngineering and Rehabilitation 10 (2013) 30. doi:10.1186/1743-0003-10-30.
URL http://jneuroengrehab.biomedcentral.com/articles/10.1186/1743-0003-10-30

[3] H. D. Taghirad, Parallel Robots, 1st Edition, CRC Press, 2013. doi:10.1201/b16096.
URL https://www.taylorfrancis.com/books/9781466555778

[4] A. Rastegarpanah, M. Saadat, A. Borboni, Parallel Robot for Lower Limb Rehabilitation Exercises, Applied Bionics and Biomechanics 2016 (2016). doi:10.1155/2016/8584735.
URL https://www.hindawi.com/journals/abb/2016/8584735/

[5] S. Briot, W. Khalil, Dynamics of Parallel Robots, 1st Edition, Vol. 35, Springer International Publishing, 2015. doi:10.1007/978-3-319-19788-3.
URL http://link.springer.com/10.1007/978-3-319-19788-3

[6] Y. Jin, H. Chanal, F. Paccot, Parallel Robots, 1st Edition, Springer London, 2015. doi:10.1007/978-1-4471-4670-4_99.
URL http://link.springer.com/10.1007/978-1-4471-4670-4_99

[7] J. A. Saglia, N. G. Tsagarakis, J. S. Dai, D. G. Caldwell, Control Strategies for Patient-Assisted Training Using the Ankle Rehabilitation Robot (ARBOT), IEEE/ASME Transactions on Mechatronics 18 (2013) 1799–1808. doi:10.1109/TMECH.2012.2214228.
URL http://ieeexplore.ieee.org/document/6296716/

[8] W. Meng, Q. Liu, Z. Zhou, Q. Ai, B. Sheng, S. S. Xie, Recent development of mechanisms and control strategies for robot-assisted lower limb rehabilitation, Mechatronics 31 (2015) 132–145. doi:10.1016/j.mechatronics.2015.04.005.
URL https://linkinghub.elsevier.com/retrieve/pii/S0957415815000501

[9] M. J. Kim, W. Lee, J. Y. Choi, G. Chung, K. L. Han, I. S. Choi, C. Ott, W. K. Chung, A passivity-based nonlinear admittance control with application to powered upper-limb control under unknown environmental interactions, IEEE/ASME Transactions on Mechatronics 24 (2019) 1473–1484. doi:10.1109/TMECH.2019.2912488.

[10] M.-S. Ju, C.-C. Lin, D.-H. Lin, I.-S. Hwang, S.-M. Chen, A rehabilitation robot with force-position hybrid fuzzy controller: hybrid fuzzy control of rehabilitation robot, IEEE Transactions on Neural Systems and Rehabilitation Engineering 13 (2005) 349–358. doi:10.1109/TNSRE.2005.847354.
URL https://ieeexplore.ieee.org/document/1506821/

[11] Y. Fan, Y. Yin, Active and Progressive Exoskeleton Rehabilitation Using Multisource Information Fusion From EMG and Force-Position EPP, IEEE Transactions on Biomedical Engineering 60 (2013) 3314–3321. doi:10.1109/TBME.2013.2267741.
URL http://ieeexplore.ieee.org/document/6529106/

[12] Q. Wu, X. Wang, B. Chen, H. Wu, Development of an RBFN-based neural-fuzzy adaptive control strategy for an upper limb rehabilitation exoskeleton, Mechatronics 53 (2018) 85–94. doi:10.1016/j.mechatronics.2018.05.014.
URL https://linkinghub.elsevier.com/retrieve/pii/S0957415818300898

[13] M. Schumacher, J. Wojtusch, P. Beckerle, O. von Stryk, An introductory review of active compliant control, Robotics and Autonomous Systems 119 (2019) 185–200. doi:10.1016/j.robot.2019.06.009.

[14] T. H. Lee, W. Liang, C. W. de Silva, K. K. Tan, Force and Position Control of Mechatronic Systems, 1st Edition, Springer International Publishing, 2021. doi:10.1007/978-3-030-52693-1.
URL http://link.springer.com/10.1007/978-3-030-52693-1

[15] J. Zhou, Z. Li, X. Li, X. Wang, R. Song, Human–Robot Cooperation Control Based on Trajectory Deformation Algorithm for a Lower Limb Rehabilitation Robot, IEEE/ASME Transactions on Mechatronics 26 (2021) 3128–3138. doi:10.1109/TMECH.2021.3053562.
URL https://ieeexplore.ieee.org/document/9335500/

[16] C. Gosselin, J. Angeles, Singularity analysis of closed-loop kinematic chains, IEEE Transactions on Robotics and Automation 6 (1990) 281–290. doi:10.1109/70.56660.
URL http://ieeexplore.ieee.org/document/56660/

[17] O. Altuzarra, C. Pinto, R. Aviles, A. Hernandez, A Practical Procedure to Analyze Singular Configurations in Closed Kinematic Chains, Vol. 20, Institute of Electrical and Electronics Engineers Inc., 2004. doi:10.1109/TRO.2004.832798.
URL http://ieeexplore.ieee.org/document/1362689/

[18] A. McDaid, Y. H. Tsoi, S. Xie, MIMO Actuator Force Control of a Parallel Robot for Ankle Rehabilitation, in: Interdisciplinary Mechatronics, 1st Edition, John Wiley & Sons, Inc., 2013, pp. 163–208. doi:10.1002/9781118577516.ch8.
URL https://onlinelibrary.wiley.com/doi/10.1002/9781118577516.ch8

[19] M. Dong, W. Fan, J. Li, X. Zhou, X. Rong, Y. Kong, Y. Zhou, A New Ankle Robotic System Enabling Whole-Stage Compliance Rehabilitation Training, IEEE/ASME Transactions on Mechatronics 26 (2021) 1490–1500. doi:10.1109/TMECH.2020.3022165.

[20] P. Araujo-Gómez, M. Díaz-Rodríguez, V. Mata, O. A. González-Estrada, Kinematic analysis and dimensional optimization of a 2R2T parallel manipulator, Journal of the Brazilian Society of Mechanical Sciences and Engineering 41 (2019) 425. doi:10.1007/s40430-019-1934-1.
URL http://link.springer.com/10.1007/s40430-019-1934-1

[21] J. L. Pulloquinga, V. Mata, Á. Valera, P. Zamora-Ortiz, M. Díaz-Rodríguez, I. Zambrano, Experimental analysis of Type II singularities and assembly change points in a 3UPS+RPU parallel robot, Mechanism and Machine Theory 158 (4 2021). doi:10.1016/j.mechmachtheory.2020.104242.

[22] J. L. Pulloquinga, R. J. Escarabajal, J. Ferrándiz, M. Vallés, V. Mata, M. Urízar, Vision-Based Hybrid Controller to Release a 4-DOF Parallel Robot from a Type II Singularity, Sensors 21 (12) (6 2021). doi:10.3390/S21124080.
URL https://www.mdpi.com/1424-8220/21/12/4080

[23] A. Morell, M. Tarokh, L. Acosta, Solving the forward kinematics problem in parallel robots using Support Vector Regression, Engineering Applications of Artificial Intelligence 26 (2013) 1698–1706. doi:10.1016/j.engappai.2013.03.011.
URL https://linkinghub.elsevier.com/retrieve/pii/S0952197613000535

[24] J. L. Pulloquinga, R. J. Escarabajal, M. Vallés, Ángel Valera, V. Mata, Trajectory Planner for Type II Singularities Avoidance Based on Output Twist Screws, in: Advances in Robot Kinematics 2022, Vol. 24, Springer International Publishing, 2022, pp. 445–452. doi:10.1007/978-3-031-08140-8_48.
URL https://link.springer.com/10.1007/978-3-031-08140-8_48

[25] J. L. Pulloquinga, R. J. Escarabajal, Ángel Valera, M. Vallés, V. Mata, A Type II singularity avoidance algorithm for parallel manipulators using output twist screws, Mechanism and Machine Theory 183 (2023) 105282. doi:10.1016/j.mechmachtheory.2023.105282.
URL https://linkinghub.elsevier.com/retrieve/pii/S0094114X23000551

[26] D. A. Neumann, Kinesiology of the Musculoskeletal System: Foundations for Rehabilitation, 3rd Edition, Elsevier Health Sciences, 2016.

[27] M. Vallés, P. Araujo-Gómez, V. Mata, A. Valera, M. Díaz-Rodríguez, Álvaro Page, N. M. Farhat, Mechatronic design, experimental setup, and control architecture design of a novel 4 DoF parallel manipulator, Mechanics Based Design of Structures and Machines 46 (2018) 425–439. doi:10.1080/15397734.2017.1355249.
URL https://doi.org/10.1080/15397734.2017.1355249

[28] P. Araujo-Gómez, V. Mata, M. Díaz-Rodríguez, A. Valera, A. Page, Design and Kinematic Analysis of a Novel 3UPS/RPU Parallel Kinematic Mechanism With 2T2R Motion for Knee Diagnosis and Rehabilitation Tasks, Journal of Mechanisms and Robotics 9 (2017) 061004. doi:10.1115/1.4037800.
URL https://doi.org/10.1115/1.4037800

[29] M. Özdemir, Removal of singularities in the inverse dynamics of parallel robots, Mechanism and Machine Theory 107 (2017) 71–86. doi:10.1016/j.mechmachtheory.2016.09.009.
URL https://linkinghub.elsevier.com/retrieve/pii/S0094114X1630235X

[30] A. Valera, M. Díaz-Rodríguez, M. Valles, E. Oliver, V. Mata, A. Page, Controller–observer design and dynamic parameter identification for model-based control of an electromechanical lower-limb rehabilitation

system, International Journal of Control 90 (2017) 702–714. doi: 10.1080/00207179.2016.1215529.
URL https://doi.org/10.1080/00207179.2016.1215529

[31] Y. Takeda, H. Funabashi, Motion Transmissibility of In-Parallel Actuated Manipulators, JSME international journal. Ser. C, Dynamics, control, robotics, design and manufacturing 38 (1995) 749–755. doi: 10.1299/jsmec1993.38.749.
URL http://www.jstage.jst.go.jp/article/jsmec1993/38/4/38_4_749/_article

[32] J. Wang, C. Wu, X. J. Liu, Performance evaluation of parallel manipulators: Motion/force transmissibility and its index, Mechanism and Machine Theory 45 (2010) 1462–1476. doi:10.1016/j.mechmachtheory.2010.05.001.
URL https://www.sciencedirect.com/science/article/pii/S0094114X10000789

[33] J. L. Pulloquinga, R. J. Escarabajal, V. Mata, A. Valera, I. Zambrano, A. Rosales, Gravitational Base Parameters Identification for a Knee Rehabilitation Parallel Robot, International Journal on Advanced Science, Engineering and Information Technology 12 (2022) 501. doi:10.18517/ijaseit.12.2.15289.
URL http://ijaseit.insightsociety.org/index.php?option=com_content&view=article&id=9&Itemid=1&article_id=15289

[34] G. N. Williams, T. Chmielewski, K. S. Rudolph, T. S. Buchanan, L. Snyder-Mackler, Dynamic Knee Stability: Current Theory and Implications for Clinicians and Scientists, Journal of Orthopaedic & Sports Physical Therapy 31 (2001) 546–566. doi:10.2519/jospt.2001.31.10.546.
URL http://www.jospt.org/doi/10.2519/jospt.2001.31.10.546

[35] M. Solomonow, M. Krogsgaard, Sensorimotor control of knee stability. A review, Scandinavian Journal of Medicine & Science in Sports 11 (2001) 64–80. doi:10.1034/j.1600-0838.2001.011002064.x.
URL http://doi.wiley.com/10.1034/j.1600-0838.2001.011002064.x

[36] V. Racic, A. Pavic, J. Brownjohn, Experimental identification and analytical modelling of human walking forces: Literature review, Journal of Sound and Vibration 326 (2009) 1–49. doi:10.1016/j.jsv.2009.04.020.
URL https://linkinghub.elsevier.com/retrieve/pii/S0022460X09003381

## Biographies

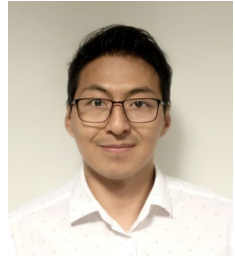

**José L Pulloquinga** received his degree in Mechatronics Engineering in 2014 from The University of the Armed Forces-ESPE, Ecuador. He received his M.Sc Degree in Mechatronics and Robotics in 2018 from the Escuela Politécnica Nacional, Ecuador. Currently, he is a Ph.D. candidate in automatic, robotics and industrial computing from Universidad Politécnica de València (UPV), Spain. His research interests include robot control, singularities, and artificial intelligence.

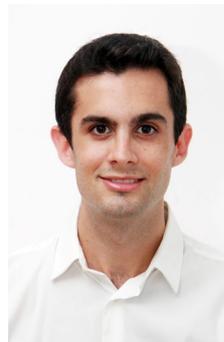

**Rafael J. Escarabajal** received a B.Eng. in Industrial Engineering in 2017 and an M.Sc Degree in Automatic Control and Robotics in 2019 from Universidad Politècnica de València (UPV), Spain. Currently, he is a Ph.D. candidate in robotics and industrial computing at the Instituto Universitario de Automática e Informática Industrial, UPV, Spain. His research interests include robot learning and control, optimisation, and artificial intelligence.

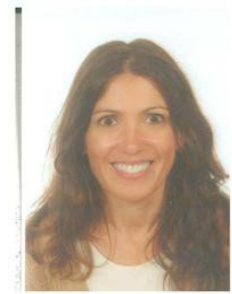

**Marina Vallés** She received a B.S. degree in computer science in 1996, an M.S. degree in CAD/CAM/CIM in 1997, and a Ph.D. degree in automation and industrial computing in 2004 from Universitat Politècnica de València, Valencia (UPV), Spain. She is currently a Full Professor in the Systems Engineering and Automation Area of the Universitat Politècnica de València. Her publications include 30 articles in journals and book chapters and 84 presentations at congresses. Her research focuses on implementing controllers in real-time and limited resource systems.

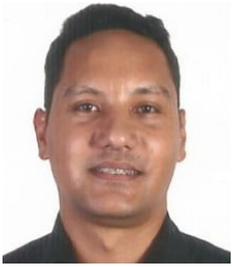
**Miguel Diaz-Rodriguez** is currently a Full Professor and Head of the School of Mechanical Engineering at the Universidad de los Andes, Venezuela. He received his Ph.D. in Mechanical Engineering from Universitat Politècnica de València in 2009 and his Mechanical engineering degree from Universidad de los Andes, Venezuela in 2000. His research interests include parallel manipulators, parameter identification and developing mechatronics systems, and dynamic modelling.

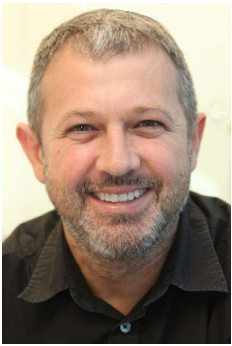
**Angel Valera** received the B.Eng. in Computer Science in 1988, the M.Sc. Degree in Computer Science in 1990, and his Ph.D. in Control Engineering in 1998 from the Universitat Politècnica de València (UPV), Spain. He has been a professor of automatic control at the UPV since 1989 and now he is a Full Professor at the Systems and Control Engineering Department. He has taken part in 75 research and mobility projects funded by local industries, the government and the European community and he has published over 190 technical papers in journals, technical conferences, and seminars. His research interests include mechatronics, industrial and mobile robots and robot control.

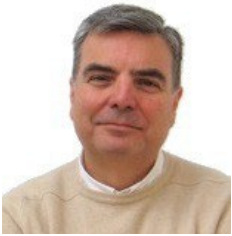
**Vicente Mata** is a Mechanical Engineer from the Escuela Técnica Superior de Ingenieros Industriales of Valencia, Spain. In 1985 he got his Ph.D. degree from the same institution. Since 2002, he has been a professor attached to the Department of Mechanical and Materials Engineering, where he teaches Theory of Machines and Mechanisms, Robotics and Mechanics. His research interests are currently focused on design of parallel robots, identification of dynamic parameters and development of biomechanical models. He has been director / co-director of 7 doctoral theses and is the author of more than 80 papers published in JCR indexed journals.